# Beta Thalassemia Carriers detection empowered federated Learning

**Muhammad Shoaib Farooq[1], Hafiz Ali Younas[1]**
[1]Department of Artificial Intelligence, School of System and Technology, University of Management and Technology, Lahore, 54000, Pakistan
Corresponding author: Muhammad Shoaib Farooq (shoaib.farooq@umt.edu.pk)

**Abstract.** Thalassemia is a group of inherited blood disorders that happen when hemoglobin, the protein in red blood cells that carries oxygen, is not made enough. It is found all over the body and is needed for survival. If both parents have thalassemia, a child's chance of getting it increases. Genetic counselling and early diagnosis are essential for treating thalassemia and stopping it from being passed on to future generations. It may be hard for healthcare professionals to differentiate between people with thalassemia carriers and those without. The current blood tests for beta thalassemia carriers are too expensive, take too long, and require too much screening equipment. The World Health Organization says there is a high death rate for people with thalassemia. Therefore, it is essential to find thalassemia carriers to act quickly. High-performance liquid chromatography (HPLC), the standard test method, has problems such as cost, time, and equipment needs. So, there must be a quick and cheap way to find people carrying the thalassemia gene. Using federated learning (FL) techniques, this study shows a new way to find people with the beta-thalassemia gene. FL allows data to be collected and processed on-site while following privacy rules, making it an excellent choice for sensitive health data. Researchers used FL to train a model for beta-thalassemia carriers by looking at the complete blood count results and red blood cell indices. The model was 92.38 % accurate at telling the difference between beta-thalassemia carriers and people who did not have the disease. The proposed FL model is better than other published methods in terms of how well it works, how reliable it is, and how private it is. This research shows a promising, quick, accurate, and low-cost way to find thalassemia carriers and opens the door for screening them on a large scale.

## 1. Introduction

Thalassemia originates from the Greek terms Thalassa, meaning "sea," and Hema, meaning "blood". The name refers to the coastal area where George Whipple first discovered the condition in 1936 [1]. Thalassemia has a significant occurrence rate in Sub-Saharan Africa, the Mediterranean, the Middle East, the Indian subcontinent, East Asia, and Southeast Asia. In these areas, it is a common genetic condition [2], [3]. Approximately 68,000 newborns with various forms of thalassemia are born each year. Beta thalassemia is particularly widespread, affecting 1.5% of the world's population, or between 80 and 90 million people. According to a report by the World Health Organization (WHO), 5.1% of people worldwide are carriers of beta thalassemia [4],[5]. A genetic condition called thalassemia is passed from one generation of parents to the next, and the likelihood of inheriting the illness rises if both parents are carriers. Severe anaemia is brought on by this disease's decrease in haemoglobin levels, which lowers the quantity of functional red blood cells [6]. Red blood cell transfusions are regularly necessary for beta-thalassemia major patients to stay alive [7]. As it transports oxygen from the lungs to all body tissues and returns carbon dioxide to the lungs, haemoglobin is one of its most important roles in the human body. Without haemoglobin, red blood cells cannot function properly. [8],[9],[10]. The severity of the condition, categorized using terms like trait, carrier, intermedia, or major, or the specific portion of haemoglobin affected by "beta" or "alpha", determine how thalassemia is classified. The two components of haemoglobin, alpha and beta, carry oxygen to every cell in the body [5]. The terms "alpha" or "beta" thalassemia denote the absence of one of these two haemoglobin components, which results in insufficient haemoglobin levels. It is known as alpha thalassemia if the alpha part is affected and beta-thalassemia if the beta part is affected [11], [9], [7].

The severity of thalassemia can be categorized using terms such as "trait," "minor," "intermedia," or "major." Individuals with thalassemia major may experience severe symptoms, including the need for frequent blood transfusions; those with thalassemia trait may show no signs or only mild anaemia. The severity can significantly affect a person's life and necessitates careful medical attention [9].

Medical professionals typically use a combination of patient clinical reports and blood smear images to diagnose the disease accurately. However, it can frequently be challenging to distinguish between various illnesses that fall under the same category based solely on clinical reports. In addition, a more thorough understanding of the patient's condition can be obtained using these diagnostic tools, enabling doctors to make an accurate and efficient diagnosis [12]. HPLC, haemoglobin electrophoresis,

PCR mutation screening, and DNA tests are current methods for screening hemoglobinopathies (such as thalassemia). Still, they are expensive and require specialized tools and qualified personnel [13].

Wide ranges of tests that can give a precise picture of the patient's condition are needed to differentiate between beta-thalassemia and iron deficiency anaemia. These exams consist of measuring the serum iron level, a complete blood count, high-performance liquid chromatography, calculating the ferritin and HBA2 levels, and measuring the iron-binding capacity. Even though these tests are essential for correctly diagnosing the condition, they can be expensive and not always readily available, making the diagnosis difficult [14]. This is due in large part to the impressive advances that have been made in The utilization of machine learning across a diverse array of scenarios, including medical diagnosis, autonomous vehicles, language translation, and more. Machine learning is an important AI subfield using self-improving computer algorithms to analyze massive amounts of data, learn new patterns, and improve. The primary application of this area of study is the resolution of complex problems for which humans can only define overarching goals but not provide detailed instructions. Machine learning algorithms, in general, tackle intricate problems by modelling data distributions in high-dimensional spaces according to statistical and probabilistic principles.

Extensive ranges of biomedical problems have responded well to treatment using multiple machine learning algorithms. Many other disease types, such as brain tumours, have also seen the development of numerous models for data analysis [15], [34], and diseases like tumours in the brain [16], [17], [18]. Chronic kidney disease [19], [20], and [21]. Lung disorders [22], and iron deficiency anaemia [23]. Additionally, the smoking problem has been addressed by implementing machine learning techniques. K-nearest neighbour [25], fuzzy logic [26], [27], Deep extreme machine learning [28], deep neural network [12], and support vector machine [24], [35], [36], [37]. Finding people with the thalassemia gene is essential for reducing the number of people with this disease. The spread of thalassemia can be partially stopped by using suitable screening methods. But figuring out who has the gene is a complicated and time-consuming process that requires trained people, and specialized equipment makes it expensive and hard to get, especially in places with less technology. Developing screening methods that are easy to use and don't cost much money is essential to stop the disease from spreading. This study trained machine learning models on a dataset of blood test results from 5066 patients, which included nine red blood cell parameters, to find a reliable way to find people who carry beta-thalassemia. Normalization was used to turn the blood test results into a scale from 0 to 5, and Federated Learning was used to evaluate the machine-learning model so that privacy was kept. The main goal of this study is to find an accurate and inexpensive way to find people who carry the beta-thalassemia gene. The study aimed to find a way to find people who carry the beta thalassemia gene accurately. It's done by making a system that uses CBC tests to do. The proposed method makes a big difference in the field because it is practical, cheap, and helps prevent disease spreading.

The rest of this paper is set up like this: Section II talks about the related work, Section III discusses the materials and methods used in this study, and Section IV discusses the proposed framework. In Section V, we talk about the study's results, and in Section VI, we wrap up the work. Because they focused on preprocessing techniques, data balancing, and different supervised and semi-supervised learning models, earlier research models were less accurate at diagnosing diseases. So, this study shows a new way to find diseases based on federated learning. This method combines the accuracy of different machine-learning algorithms while keeping a high level of accuracy. The primary strategy used in this study for making data accessible, completing data collection more manageable, and ensuring data security was using the Federated Learning (FL) approach. The researchers trained three different machine-learning models on their computers and then sent them to a central server for more training. This made it possible to create a global data model to find people with beta-thalassemia accurately. As a result, researchers were able to make an accurate and reliable model for figuring out who has beta-thalassemia without putting patient privacy or data security at risk.

In short, the goal of the study was to make a computer system that can diagnose thalassemia more accurately, with better results, and for less money. The proposed method, which is based on Federated Learning, is a fast, accurate, and cost-effective way to find people with beta-thalassemia. It also improves data privacy, accessibility, and security.

## 2. Related Work

Diagnosing a thalassemia carrier can be difficult, which can cause delays in effective management. Therefore, the best strategy is to use medical expertise based on patient data. Although it is expensive and out of reach for many people in developing countries, It is imperative to create an automated prediction system for identifying Thalassemia carriers to prevent the transmission of the disease to future generations. The importance of choosing the suitable model for Thalassemia carrier prediction is highlighted by the fact that the effectiveness of machine learning models varies depending on the field to which they are applied. [31]. Finding early signs of Thalassemia disease is the research goal [8]. Scientists worldwide have started looking into various technological solutions to enhance healthcare delivery by utilizing the IoT's potential to complement current services. This study thoroughly analyses several IoT-based healthcare technologies, including a survey of the platforms and network architectures used in healthcare that facilitate access to the IoT backbone and the transmission and reception of medical data. [29], [30].

Red blood cells (RBCs), notably Target Cells and Elliptocytes, can now be categorized, counted, and graded using a device. This approach is comparable to the haematologists' conventional manual practice. However, the RBC morphology grading-based disease identification is not considered the final diagnosis, and the doctor's experience is necessary for interpreting the results. The peripheral blood smear (PBS) result can be used as a foundation for haematological diagnosis by acting as an early indicator of disease. In conclusion, the created device offers a simple and trustworthy method for analyzing RBCs, which might be helpful for nearby diagnostic labs [24].

To differentiate between Iron Deficiency Anemia (IDA) and beta-thalassemia, two machine-learning techniques, namely Support Vector Machine and K-Nearest Neighbor, were employed in a decision-based diagnosis. The diagnostic model included several CBC-related parameters, and adding Red Blood Cell (RBC) indices increased its effectiveness. The diagnostic system could, however, become overly complex as the number of features rises. These results show that IDA and beta-thalassemia can be distinguished more accurately using RBC indices and machine learning techniques. Future studies could examine the use of extra diagnostic tools that could improve the system's efficiency. Edouard de Dreuzy et al. [7] cite three possible approaches that might be used to deal with the current problem. The first strategy involves using pharmacological substances to reactivate the fetal globin gene. With this approach, the medications must be given to patients continuously. Allogeneic hematopoietic stem cell transplantation is the second technique (HSCT). Gene therapy is the final strategy.

It used a decision-based system that distinguished between beta-thalassemia carriers, healthy people operating decision trees, ANN, and a naive Bayes classifier. The dataset was gathered at Chandigarh, India's postgraduate medical education and research institute. It was determined that both ratings were entirely sensitive. The combined screening score for BTT and HbE was 91.74 percent and 79.25 percent, respectively, for thalassemia characteristics (BTT). Although the mechanism distinguishes between two major haemoglobin variants, it still needs to be validated using datasets gathered from various countries for use and unification.

[6] A prediction model was created using supervised machine learning techniques to determine the risk of thalassemia disease. The Waikato Environment for Knowledge Analysis (WEKA) tool was used to analyze the data gathered from questionnaires and medical staff. In addition to clinical factors like spleen enlargement, family history, urine colour, diabetes, and inherited disease status, the model used demographic variables like age, marital status, gender, social class, and ethnicity for identification. Multi-layer perception (MLP) and Nave Bayes classifier were used as models, with the dataset consisting of 57% disease carriers and 43% non-carriers. The study discovered that MLP was a more efficient and reliable method for determining patients in Nigeria who were at risk for thalassemia. This research may help with early detection and treatment, thalassemia prevention in the next generation, and lessening the financial burden of treatment on developing nations.

Zaylaa et al. [3] The average Intersection over Union (IoU) score for predicting thalassemia with the aid of Photothermal Tomography Analysis Prediction Time Augmentation (PTA) was 88%. In contrast, it was 82% without PTA, according to quantitative findings from the study. The study also developed a new loss score metric and discovered that as this score falls, the accuracy of thalassemia prediction rises. According to these results, PTA can significantly increase the precision of thalassemia diagnosis, and the new combined loss score metric can offer a more accurate predictor of disease risk. This knowledge may help create Thalassemia diagnostic tools that are more accurate and enhance patient outcomes.

Azam et al. [27] assert that this study introduced a novel method for classifying celiac sickness by means of the IT2FLS type-2 fuzzy system. Based on the proficiency and knowledge of pathologists and gastroenterologists, specific histological characteristics that significantly influence the disease's severity were identified to determine the system's input variables. In addition, the Fuzzy C-means clustering algorithm improved the performance of IT2FLS in computing the center of membership functions as specified by the fuzzy rules. The objective of developing a model based on fuzzy logic was to classify thalassemia diseases Susanto et al. [32].

Sadiq et al. [33] To correctly identify patients with thalassemia from their Complete blood count (CBC) test data, an ensemble classifier model was built in this study using a random forest, support vector machine, and a gradient boosting machine. A dataset of CBC reports from the record of the patients who took from the Punjab thalassemia prevention programme served as the model's testing ground (PTPP). This research utilised the input parameters for red blood cells, such as haemoglobin, hematocrit, mean cell volume, mean cell hemoglobin concentration, mean cell hemoglobin, RBC distribution width, platelet count, and white blood cells. The study's conclusion states that the ensemble classifier model detected beta-thalassemia carriers with a 93% accuracy rate.

Qais Mohammed and Jamal Mustafa [34] presented various machine-learning techniques and artificial intelligence-based methods by analyzing the CBC test variables, including RBC, HGB, MCV, HTC, and HB, to classify and detect thalassemia. Identifying patients with minor thalassemia alpha and major thalassemia beta was the primary goal of this system. This study used decision trees, Naive Bayes, and support vector machines as classification techniques.

Afifah et al. [35] explain how the accuracy of thalassemia data classification was improved by employing hyperparameter optimization via Grid Search on a Support Vector Machine with an RBF kernel. After conducting various experiments, the SVM model's optimal parameters for holdout validation were determined as $C = 428.13$ and $y = 1.83 * 10 - 5$. This configuration yielded a 100% accuracy rate on 90% of the training data. For k-fold cross-validation, the ideal parameters were $C = 4832.93$ and $y = 1.83 * 10 - 5$, resulting in a maximum accuracy of 100% with $k = 10$. In the absence of hyperparameter optimization, the SVM model with RBF kernel defaulted to $C = 1$ and scale, which produced an accuracy of 73.33% for holdout validation and 57.14% for 10-fold cross-validation. The proposed system, which uses a Support Vector Machine (SVM) classifier, identified seven types of red blood cells, including standard, schistocytes, elliptocytes, dacrocytes, spherocytes, target cells, and stomatocytes, with an accuracy of 93.33%. It also successfully identified cells that were not previously known, purposed by Carlos et al. [36]. Early screening for suspected thalassemia patients is required to stop the disease from passing to future generations. The three kernels of the Twin Support Vector Machine classifier, as described by Andhika et al. [37], were found to be effective in this study's implementation of TSVM to classify data related to thalassemia.

## 3. Method and Material

This section of the paper will review the preprocessing procedures, the machine-learning framework used, and an overview of the dataset used to train the machine to predict beta-thalassemia carriers.

### 3.1 Description of Dataset

The Punjab Thalassemia Prevention and Programme (PTPP) compiled the dataset for this research. The Pakistan Thalassemia Prevention Program (PTPP) is a public sector research institute dedicated to preventing thalassemia and other inherited disorders in Pakistan. The Department of Specialized Healthcare and Medical Education administers it. To achieve its primary aim of identifying thalassemia carriers, the program conducts annual screenings on more than 300,000 patients. While the PTPP also provides medical care for patients with thalassemia major, it emphasizes the significance of screening the extended family to reduce the number of carriers. To achieve this, if one member of a family is diagnosed with thalassemia major, all of their other relatives should also be evaluated.

As shown in Figure 1, the dataset for this investigation consists of information from 5066 patients who underwent screening in 2019. Out of these patients, 3051 did not carry the thalassemia gene, while 2015 did. In addition, the data set is divided by gender and age, with 53% males and 47% females, and by age, with 54% adults and 46% children.

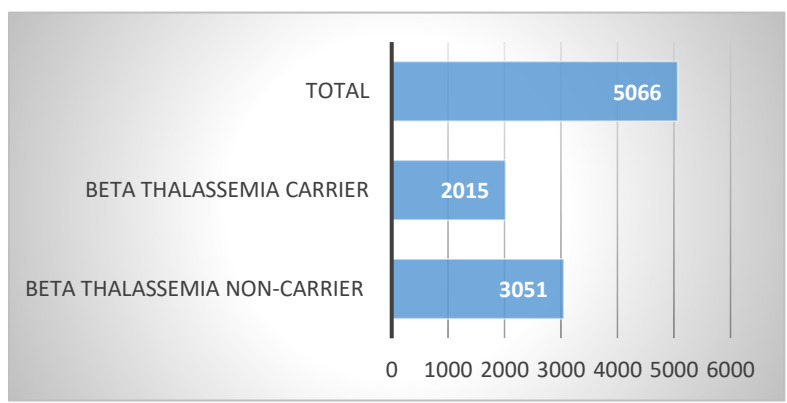

**Fig. 1**. The distribution of the dataset is shown based on its class categories.

### 3.2 Features

The attributes used in the research were complete blood count (CBC) parameters and individual demographic data. Red blood cell count (RBC), haemoglobin (HB), hematocrit (HCT), mean corpuscular volume (MCV), mean corpuscular hemoglobin (MCH), mean corpuscular haemoglobin concentration (MCHC), red cell distribution width (RDW), platelet count (PLT), and white blood cell count are the components of a complete blood count (CBC). The demographic data contains gender and age. (WBC). Table 1 provides a thorough description of these traits.

**Table 1**. Features description were used in the training

| ATTRIBUTE | DESCRIPTION | NORMAL RANGE | MEASUREMENT UNIT | EXPLANATION |
|---|---|---|---|---|
| **RBC** | Red blood cell count | 4-5 x 1012 cells/l | Cells/liter | Represents the number of red blood cells in a sample of blood. |
| **HB** | Haemoglobin | 13-17 g/dl (males), 12-15 g/dl (females) | G/dl | Refers to the concentration of hemoglobin molecules in the blood. |
| **HCT** | Hematocrit | 45-50% (males), 30-45% (females) | % | Indicates the percentage of blood volume made up of red blood cells. |

| **MCV** | Mean cell volume | 80-100 fl | Hectoliters | Expresses the mean corpuscular volume (MCV) which represents the average size of red blood cells in a blood sample. |
|---|---|---|---|---|
| **MCH** | Mean cell haemoglobin | 27-34 pg/cell | Pictograms/cell | Reflects the average amount of haemoglobin per red blood cell in a sample. |
| **MCHC** | Mean cell haemoglobin concentration | 32-36% | % | Shows the average haemoglobin concentration in a volume of blood. |
| **RDW** | RBC distribution width | 11-15% | % | Demonstrates the variation in the size and shape of red blood cells. |
| **PLT** | Platelets count | 150-350 x 109/l | ×109/l | Describes the number of platelets in a sample of blood. |
| **WBC** | White blood cell count | 4-10 x 109/l | ×109/l | Specifies the number of white blood cells in a sample of blood. |
| **EXPLANATION** | The explanation for the parameter | N/a | N/a | Adds additional information or clarifications for each parameter in the table |

This study uses a scatter plot matrix to look at how the dataset's characteristics relate to one another. The data elements are colour-coded according to their class, defined in the Data Frame's "Class" column, as seen in Figure 2. The x-axis and y-axis represent two distinct features in each plot of the matrix, giving information about their relationship. The hue parameter separates the data points based on their class. The diagonal plots also show how each characteristic is distributed. In the matrix, each scatter plot can be seen twice, once in the upper half and once in the bottom half. Through the identification of feature correlations and possible outliers using this visualization, the dataset can be fully understood. Better analysis is made possible by understanding the relationships between the different features and the target variable.

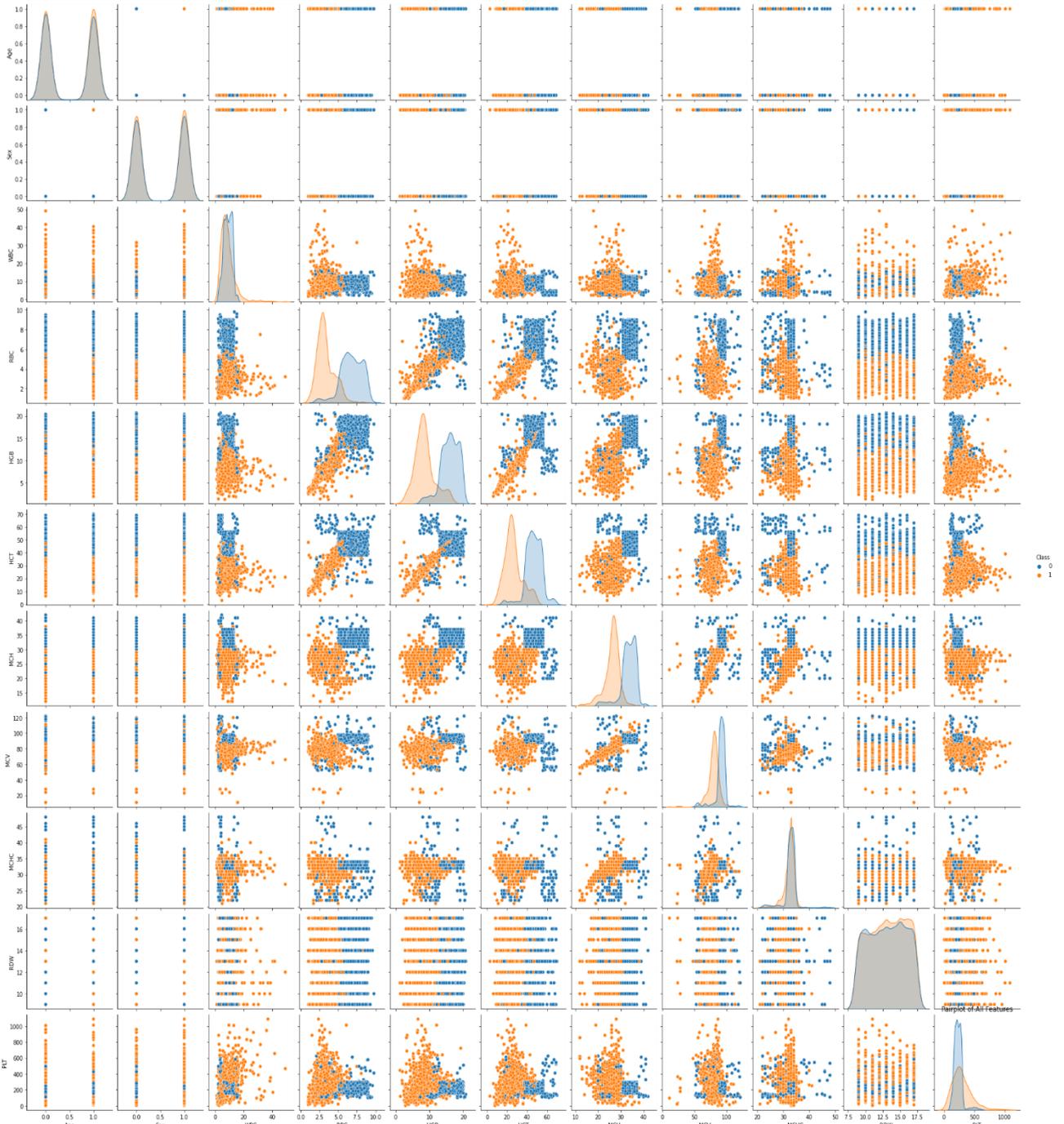

**Fig. 2.** The distribution of the dataset shown based on its class categories

### 3.3. Data Cleaning
"Data cleaning" is one of the crucial phases of performing research. This process ensures the accuracy and dependability of the dataset. However, the study's validity can be negatively impacted by incorrect, insufficient, or inconsistent data, which can result in the drawing of incorrect conclusions [38].

- ➢ Ensure that the dataset accurately represents the data that is being researched by addressing any missing values that may exist within it.
- ➢ When the missing value could not be located in the health report, the corresponding record was removed from the dataset to ensure that the data remained accurate and would not suffer corruption.
- ➢ We removed from the dataset any attributes deemed insignificant, such as the patient's name, the test date, their name, address, and mobile number.

- The age values contained in the dataset were normalized into two categories: 0 represented children (patient age less than 18), and 1 designated adult (patient age greater than 18), with the range of values falling between 0 and 1.
- The gender attribute in the dataset was normalized to a binary value, where 0 denoted female and 1 denoted male. The binary value was then converted to a string.
- Regarding the evaluations included in the dataset, each review was standardized into six categories ranging from 0 to 5 points. A test result lower than the normal range was represented by the value 0, while a test result higher than the normal range was represented by the value 5. Within the boundaries of the normal range, the values 1, 2, 3, and 4 denoted four equal divisions.

The resultant class in the dataset was denoted by a binary value of either 0 or 1, depending on the case. A value of 0 indicated that the individual did not carry the beta-thalassemia trait, and a value of 1 signified that the individual exhibited the characteristic.

## 4. Proposed Framework

A promising method for identifying beta-thalassemia carriers has been included in the suggested model through the integration of a Decision Tree, Naive Bayes, and Support Vector Machine learning techniques. The datasets were initially trained locally, arranged by features, and sent to a central server for aggregation via Federated Learning to guarantee patient data safety and security. As a result, a final global model has been developed, increasing model accuracy and decreasing the expense of screening. The suggested FL model's three-layer architecture is shown in Fig. 1, which offers a thorough approach to identifying beta-thalassemia carriers while upholding patient privacy. Patient data is gathered and preprocessed in the first layer and then trained on the data in the second layer by various machine learning algorithms. Finally, in the third layer, a global model for identifying beta-thalassemia carriers is developed using the findings from the individual models. Overall, this FL model offers a fresh and effective way to identify beta-thalassemia bearers while protecting patient information.

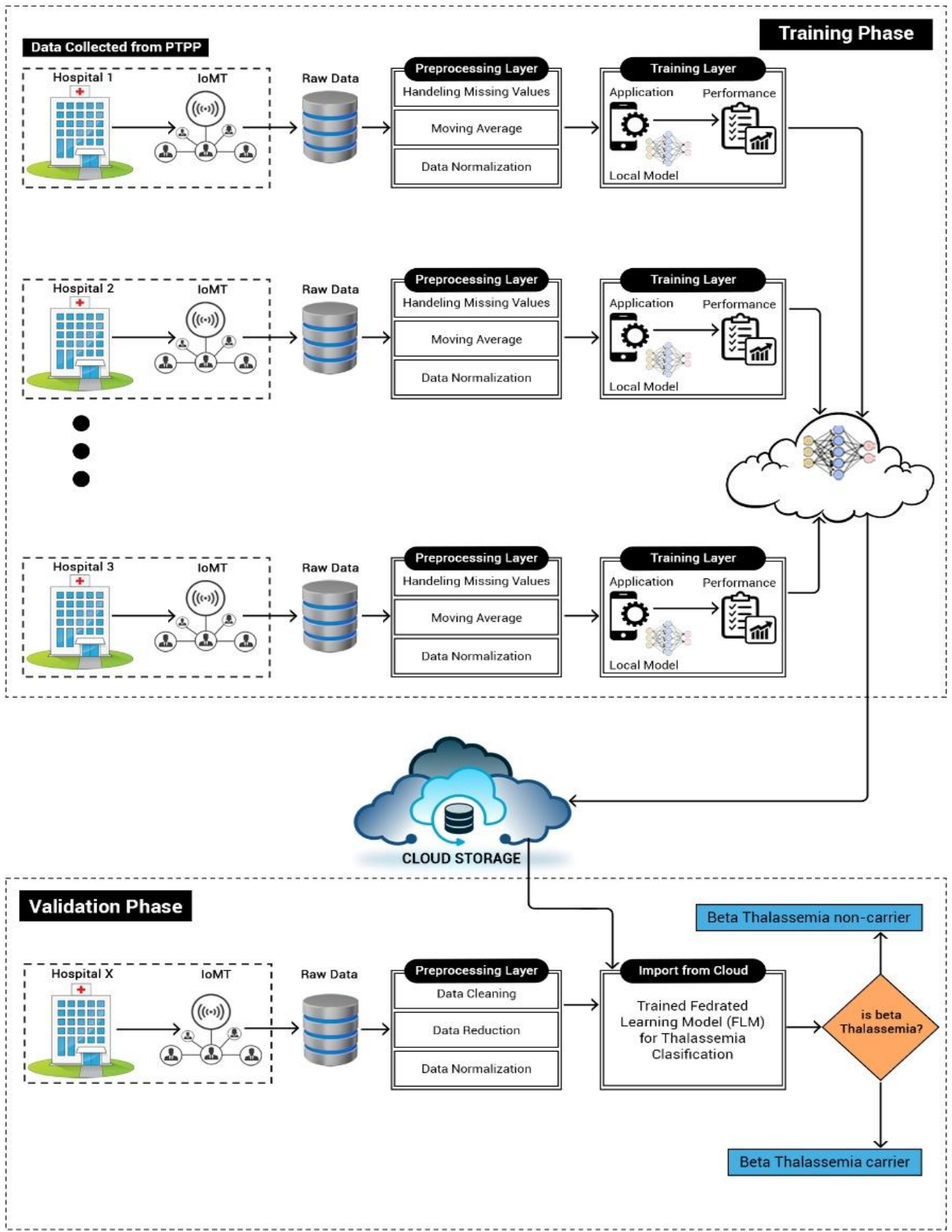

**Fig.3.** Proposed federated learning-based model for beta thalassemia carriers identification.

## 4.1. Hierarchical Architecture of Layer

### 4.1.1. Data Acquisition Layers
As the component responsible for collecting raw data, the data acquisition layer is important in the beta-thalassemia carrier identification system. In addition, Internet of Medical Things (IoMT) devices, wearables, and edge devices can gather and send real-time data about a patient's health.

Beta-thalassemia can be diagnosed using a variety of traits found in the data acquired by the data acquisition layer. Clinical parameters such as age, gender, blood test results, and others can be used to identify patients with beta-thalassemia carriers or at risk for getting the disease.

The gathered data is stored in the system database to be processed and analyzed later, guaranteeing its accuracy and completeness. The subsequent layers of the beta-thalassemia carrier detection system rely on the accurate and complete data acquired by the data acquisition layer to identify and diagnose beta-thalassemia carriers correctly; failure to do so could have serious consequences for patient health and outcomes.

Because of this, the data acquisition component is crucial to the beta-thalassemia carrier detection system's overall performance. Furthermore, as the data gathered here forms the basis for the rest of the system, its accuracy and efficiency are critical to its success in detecting and diagnosing beta-thalassemia carriers.

### 4.1.2. Preprocessing Layer
The preprocessing layer makes up the second component of the suggested beta-thalassemia detection model. Before sending the data to the machine learning algorithms for additional processing, its primary goal is to manage missing or noisy data values. The problem of missing data is dealt with using various methods by the preprocessing layer, which is in the center of the model. Among these techniques are moving averages, mean/median/mode imputation, regression imputation, numerous imputations, and k-nearest neighbour imputation. For instance, the average moving method was applied to the electronic medical record (EMR) dataset, and missing values were replaced with the average of the observations that came before and after them.

In order to eliminate noisy values and make sure that the data falls within the parameters of a specified range, the preprocessing layer is also in charge of normalizing the data. As an illustration, the normalization process divides blood test results into six divisions, transforming the real values into a range of 0 to 5. This layer's goal is to raise the data's dependability and quality so that machine learning algorithms can generate more accurate results.

The beta-thalassemia carrier detection system total effectiveness depends on the preprocessing layer's functionality. Accurate and comprehensive data are necessary for machine learning to generate reliable results. This layer improves the dataset's precision, making recognising beta-thalassemia carriers more straightforward and developing more accurate forecasts. Therefore, the preprocessing layer plays a crucial role in the beta-thalassemia carrier detection model to guarantee that the dataset is prepared for machine learning algorithms.

### 4.1.3. Training Layers
There are four machine learning models were used in the third layer of the beta-thalassemia carrier identification system, one global model and three local models. This layer is the most important component of the system. Due to the fact that the machine learning models used to detect and diagnose beta-thalassemia carriers are taught in this layer, it is also referred to as the training layer.

The second layer of preprocessing deals with missing or noisy values before the data is fed into the machine learning algorithms. As a result, predictions are more precise because the data is trustworthy and accurate. The local models are then trained using a variety of machine learning methods, including decision trees, naive Bayes, and support vector machines, on the preprocessed data. The preprocessed data is categorized into pertinent groups during the training phase.

Following training, the local models' performance is assessed using measures like accuracy, sensitivity, and specificity. The site is then represented in the global model training procedure by the model that performs the best. Finally, the federated averaging (FedAvg) algorithm, a distributed machine learning technique that allows multiple sites to train a global model without sharing their local data, is used to train the global model. The FedAvg algorithm combines the local models' parameters to create a global model that can make precise forecasts based on new data.

Transfer learning is used in addition to the FedAvg algorithm to enhance the model's overall efficiency. When using knowledge from other datasets, particularly when the dataset is small, the model can increase the accuracy of its predictions. Transfer learning is a machine learning technique that transfers knowledge acquired from one task to another.

Overall, training the machine learning models that will be used to discover beta-thalassemia carriers depends heavily on the third layer of the beta-thalassemia carrier detection system. While still producing precise and reliable predictions, using global and local models and strategies like federated learning and transfer learning ensures that the data stays private and secure.

### 4.2. Training of all Models

#### 4.2.1. Decision Tree (Local Model 1)
The decision tree algorithm is frequently used in medical research for classification and prediction assignments [39]. The suggested framework for identifying beta-thalassemia carriers has also used the decision tree algorithm. This algorithm is a great option because it can handle complex data processing and incorporate a variety of clinical features and parameters into its predictions. Each node represents a decision and its effect, and the route from the root node to the leaf nodes illustrates the sequence of decisions made to conclude. This creates a tree-like model that shows the options available and their possible outcomes. Medical professionals can benefit significantly from the decision tree algorithm in cases where disease may be caused by a number of different variables, making a diagnosis difficult. As a result, in the suggested framework, it is a useful instrument for predicting beta-thalassemia carriers.

Only 1690 of the total 5066 records were needed for training the Decision Tree method, and the resulting model is displayed in PNG format in Fig. 4. The decision tree model's visualization offers helpful insights into how it functions internally, enabling us to pinpoint the crucial characteristics that are used to make forecasts. It is also a useful algorithm for optimizing the model's hyperparameters and analyzing possible problems with overfitting or underfitting. This visualization is essential for improving the classifier model's general performance and ensuring its accuracy and dependability in practical applications.

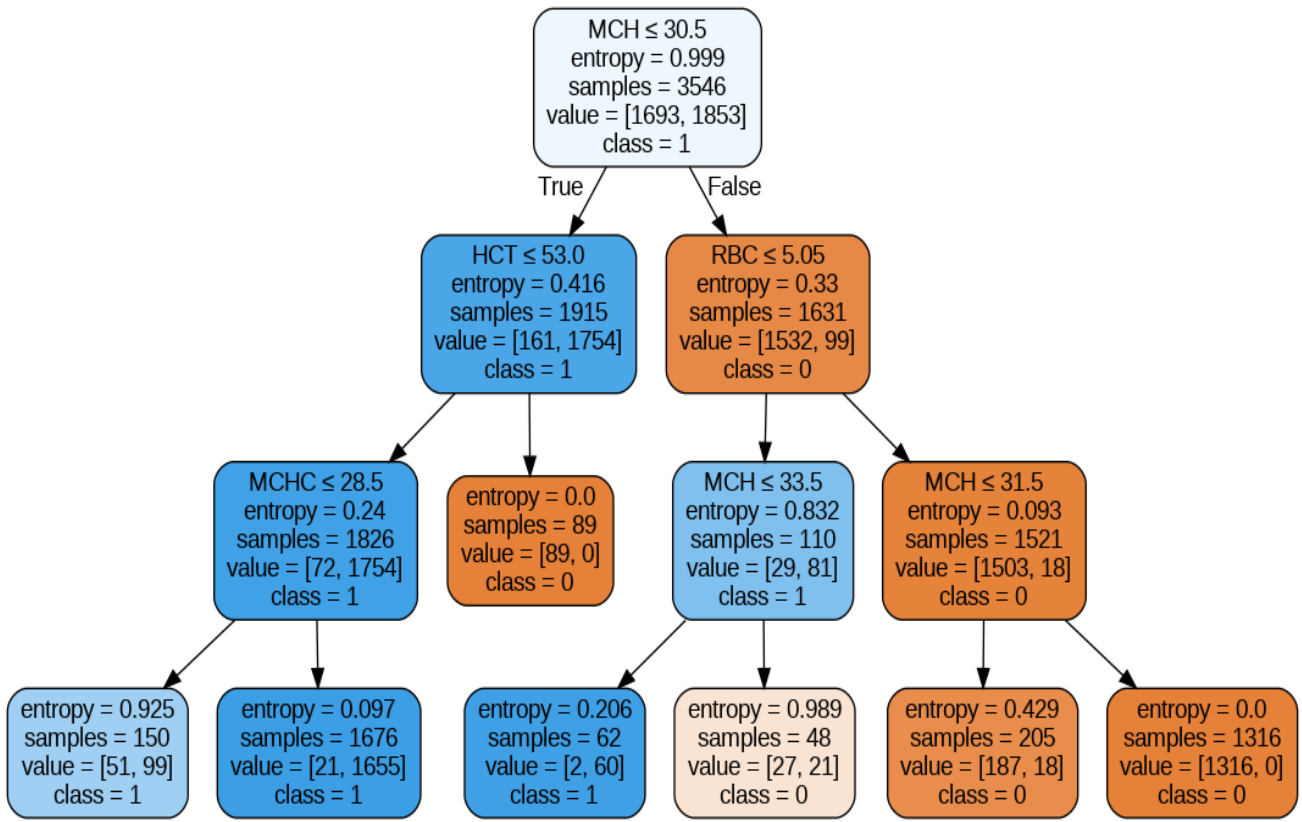

**Fig.5.** Decision tree simple model visualization

$$D(X) = -\sum Prob\ (rv) * log2\ (\ Prob\ (rv))$$

#### 4.2.2. Naïve Bayes (Local Model 2)
The Naive Bayes algorithm is frequently used in the field of health science due to its efficiency in handling large datasets with numerous variables, its ability to manage missing data, and its solid theoretical foundation. It has also demonstrated efficacy in a variety of medical applications [40]. Based on Bayes' theorem and the assumption of feature independence, this algorithm can predict the likelihood of an event based on the presence or absence of distinct features. The Naive Bayes algorithm is particularly valuable for predicting the likelihood of beta-thalassemia based on clinical data for medical professionals.

The proposed framework for beta-thalassemia carrier prediction can combine Naive Bayes with other machine learning methods, such as decision trees, to increase the accuracy of predictions. The projection of high-dimensional data onto a two-dimensional subspace using Principal Component Analysis (PCA) is depicted in Figure 5. Based on the classification model, a Naive Bayes classifier is then applied to these points with a decision boundary using a mesh grid to classify the data points into two categories. The scatter plot is depicted with the decision boundary and classification regions using a colour map, enabling visualization of the classification performance of the Naive Bayes model on the two-dimensional subspace. The Naive Bayes model was trained on a distinct set of 1689 samples than the decision tree model.

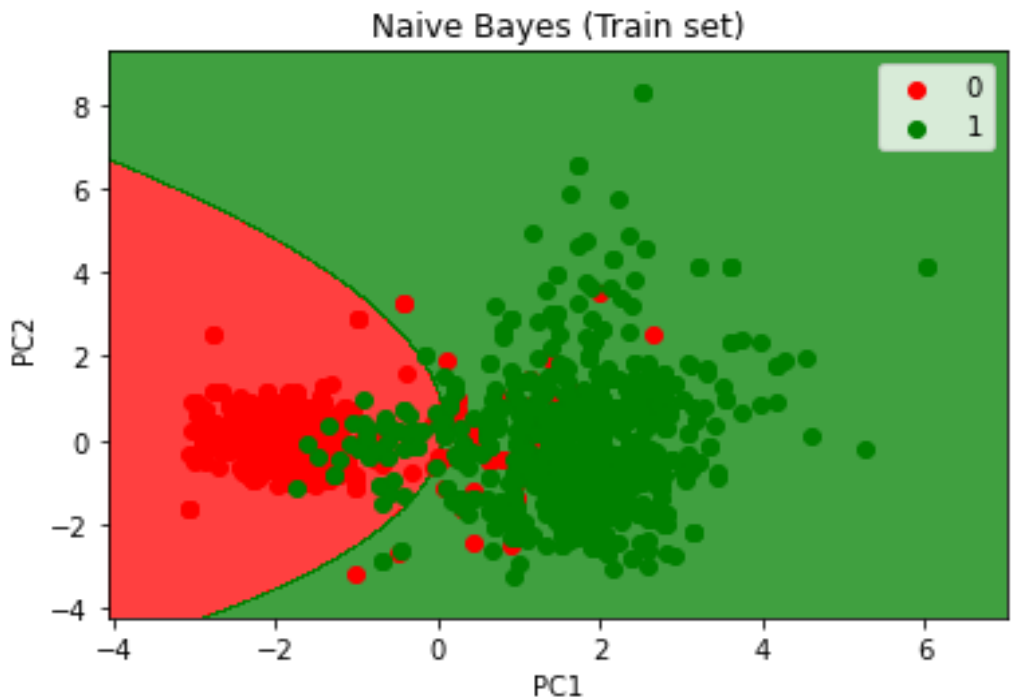

**Fig. 4.** Training set of the naïve bayes

$$Prob\ (X/Y) \ = \ Prob\ (Y/X)\ * \ (Prob\ (Y))/(Prob\ (X))$$

### 4.2.3. Support Vector Machine (Local Model 3)

Support Vector Machines (SVMs) are sophisticated classification algorithms with a broad range of applications that incorporate elements from earlier techniques. Similar to discriminant analysis, SVMs work under the premise that, provided the data is "separable," categories of the data can be created using a functional separator [41]. The Support Vector Machine (SVM) algorithm, a supervised machine learning method, is used in the proposed beta-thalassemia carrier prediction framework to find the hyperplane that offers the most significant degree of separation between data points in different categories. Because it can differentiate between people with the condition and those without it using a broad range of clinical traits, the SVM algorithm is helpful for predicting beta-thalassemia carriers. The hyperplane that best divides the groups is found after the input data are transformed into a high-dimensional space. In addition, SVM has been discovered to be helpful when a variety of potential variables cause a disease's presence. As a result, incorporating SVM into the suggested framework has the potential to deliver accurate and trustworthy predictions of beta-thalassemia carriers, which can help medical professionals make wise patient care choices.

To help us understand the data, Figures 6 and 7 show scatter plots that show the distribution of the training and testing sets, respectively. The x-axis and y-axis of the scatter plot denote the values of two characteristics, and the hue of each point denotes the class to which it belongs. This visualisation can better understand the connection and degree of separation between the features and the class labels. After the dataset was split into training and testing sets, each comprising roughly 1689 samples, the SVM model was trained on the remaining dataset. The model performs better on the test set due to this strategy, which increases its capacity for learning and generalization. Selecting the most relevant features to the problem at hand is essential for creating an accurate and dependable model.

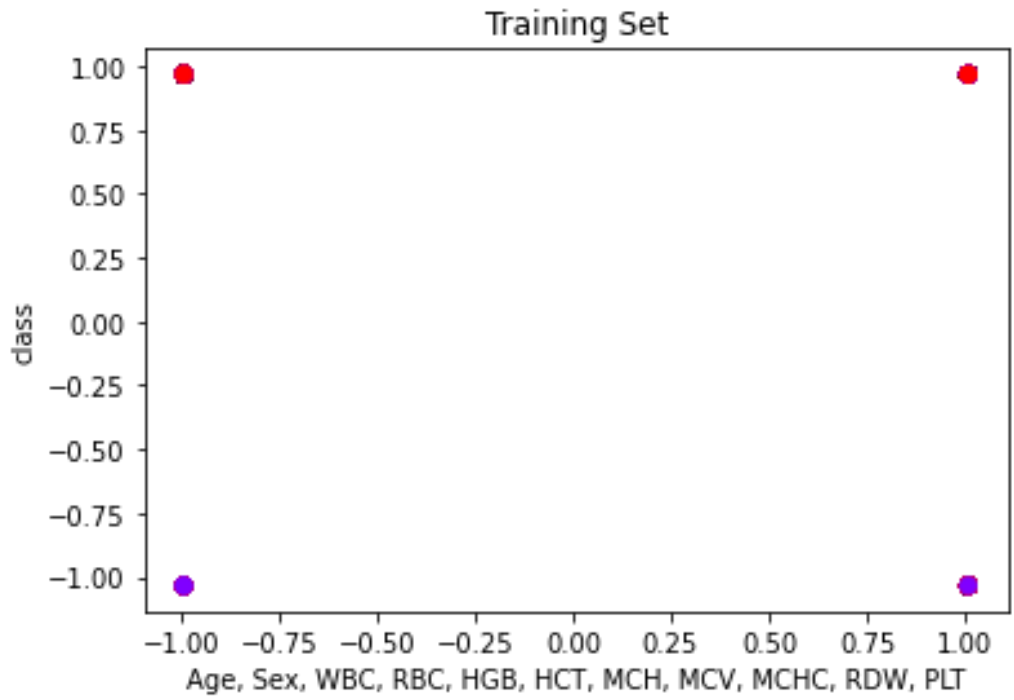

**Fig. 6.** Training of the SVM

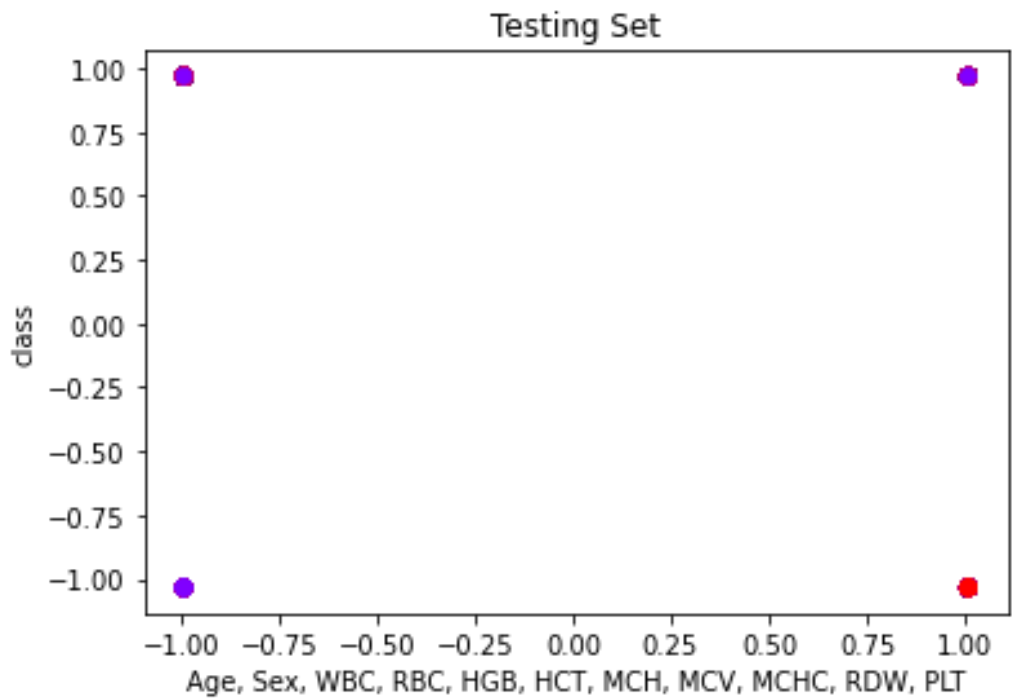

**Fig. 7.** Testing of the SVM

| Predictive Model | Advantage | Norms and/or Boundaries |
|---|---|---|
| **Decision Tree** | Decision trees are easy to understand and interpret, even for non-experts in machine learning. | Classes must be mutually exclusive, meaning that each instance must belong to only one class. |
| | The training algorithm for decision trees is efficient, making it useful for large datasets. | Decision trees rely heavily on the order of attribute selection during the tree construction process. |
| | The order of instances during the training process does not affect the final tree structure. | Missing values of an attribute can create confusion and negatively affect the performance of the model. |
| | The algorithm is highly efficient during training. | |
| **Naïve Bayes** | Naive Bayes is often used in text classification and spam filtering applications. | Naive Bayes is susceptible to the "zero-frequency" problem, where an attribute may not appear in the training data for a particular class. |
| | A probabilistic classifier calculates the probability of each class based on the given attributes. | The Laplace estimator can be used to solve the zero-frequency problem by adding a small constant to the frequency count of each attribute. |
| | Rapid and effective training algorithm. | Assumes that attributes are statistically independent of one another. |
| | Training is not affected by the order of instances. | Requires classes to be mutually exclusive. |
| **Support Vector Machine** | Can handle high-dimensional data effectively. | May suffer from overfitting if not properly regularized. |
| | Good performance on both small and large datasets. | Requires careful selection of kernel function and parameters. |
| | Robust to noise and irrelevant features. | May not perform well on imbalanced datasets. |
| | Can handle missing data and outliers. | The interpretability of the model may be limited. |
| | Can capture complex relationships between features. | The complexity of the algorithm structure can make it difficult to interpret or understand. |

**TABLE 3.** Advantages and limitations of machine learning methodologies

### 4.2.4. Federated (Global Model)

The healthcare sector has been paying close attention to the machine learning method known as federated learning. It has shown to be especially helpful for quick replies in the medical industry because it enables training machine learning models without sharing private patient information. Data security and privacy are major concerns in the healthcare industry, and federated learning ensures that patient privacy is upheld while allowing healthcare workers to create and enhance machine learning models by keeping the data on local devices [42], [43], [44].

To train a global model, federated learning entails sending locally created data models from individual devices to a central server. A client-server design is used to accomplish this, guaranteeing that the confidentiality of local data models is not jeopardized. As a result, Federated learning has many advantages, such as preventing network overload, resolving latency problems, and guaranteeing the confidentiality and anonymity of transmitted data.

The approach mentioned above was used to create a system for the prediction of beta-thalassemia carriers to show the value of federated learning in the healthcare industry. Three local models—decision trees, naive Bayes, and support vector machines—were SVM used to create a global model. The naive Bayes algorithm predicts an event's probability based on the existence or absence of various features. In contrast, the decision tree algorithm builds a tree-like model of decisions and possible outcomes based on input data. The SVM algorithm allows for the classification of new data points by dividing various groups of data into boundaries or hyperplanes. The suggested framework for beta thalassemia carrier prediction can precisely predict a person's carrier status by utilizing these machine learning techniques.

*Training steps of Global Model*
1. Obtain local models from different clients
2. Initialize a global model as a Support Vector Machine with a linear kernel
3. Extract Support Vector Machine (SVM) models from local models and concatenate their support vectors and dual coefficients
4. Compute the mean dual coefficients and set them to the global model

5. Extract Decision Tree (DT) models from local models and concatenate their feature importance
6. Compute the mean feature importance and set them to the global model
7. Set the kernel and gamma (for SVM models) or criterion and max depth (for DT models) of the global model to that of the first model of their respective types
8. Concatenate the datasets from different clients to create a global dataset
9. Train the global model with the global dataset
10. Load the test data
11. Make predictions with the global model
12. Evaluate the accuracy of the global model
13. Print the accuracy of the global mode

## 5. Result and Simulation

Python, a well-known programming language, is used to apply the suggested model for detecting beta-thalassemia carriers. The model uses federated learning (FL), a machine learning method that analyzes data with 5066 features by combining several machine learning methods based on the Support Vector Machine (SVM).

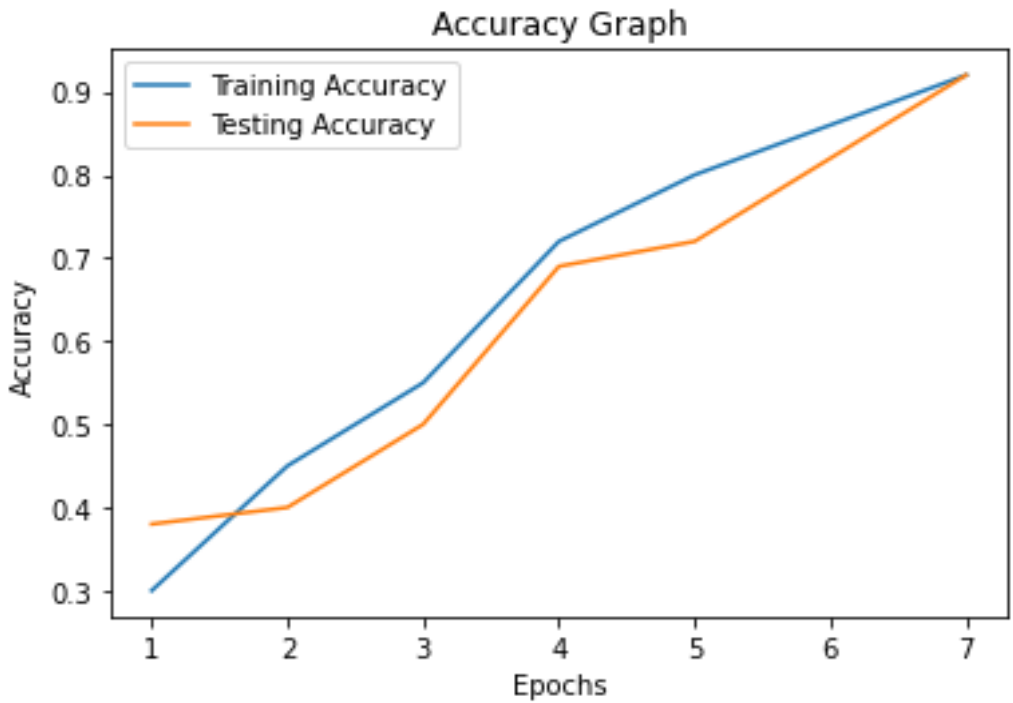

**Fig. 8.** Training and Validation of Global Model

Figure 8 shows the training and validation accuracy of the FL-based global model for detecting the beta-thalassemia carrier. The performance of the proposed training and validation global model, which was trained using 70% of the available data, is shown in this graph. The results demonstrated the model's impressive accuracy of 97.38% on the training data, demonstrating that it could successfully learn and generalize. Additionally, the global model's validation accuracy was 92.38%, showing that it did well on data it had never seen before. These encouraging findings imply that the suggested global model can successfully learn from the available data and generalize to newly collected data.

**TABLE 2.** Proposed Global Model-based Confusion Matrix (Validation)

| Approaches | Accuracy | Miss Rate |
|---|---|---|
| Global Model | 97.89% | 7.62% |
| Average Performance Proposed Model | 92.38% | 8.11% |

The validation step of the global model-based thalassemia disease identification model is shown in Table 2. This model used 5066 samples, including 3051 non-thalassemia carriers and 2015 carriers. 17 thalassemia non-carrier samples and 23 thalassemia carrier samples were forecasted by the model incorrectly, though. These results show that the suggested global model can predict thalassemia disease carrier status accurately

## 6. Conclusion

The study aims to develop a technique that can correctly assess beta-thalassemia carrier patients while maintaining the privacy of their personal data. Both medical experts and laypeople can use the federated learning-based model that has been suggested because it offers a simple, understandable methodology. Furthermore, anyone can assess their thalassemia risk by giving the system with the necessary input data.

The FL-based model that has been suggested has achieved an incredibly high degree of total precision—92.38 percent—which is impressive given the study's aim to investigate thalassemia from all of its possible perspectives. However, the potential of the currently proposed structure can be further increased by including additional forms of thalassemia, such as Alpha Max and Min and Beta Max and Min, which could enhance the effectiveness of the system.

One of the main advantages of using a federated learning approach is the provision of privacy. It is not necessary to disclose patient information to other businesses or people because it is held safely within the local data centres of each participating institution. This approach strengthens data security and confidentiality, two factors that are crucial in the medical industry and safeguard the privacy of patient information.

Overall, the study's results show a hopeful technique for identifying thalassemia carriers, which may help slow the spread of the condition and improve the quality of life for those affected. The proposed FL-based model, along with any possible extensions in the future, has the potential to become an essential tool for the diagnosis and management of thalassemia in the medical field.